\pdfoutput=1

\documentclass[11pt]{article}

\usepackage[]{ACL2023}

\usepackage{times}
\usepackage{latexsym}
\usepackage[T1]{fontenc}

\usepackage[utf8]{inputenc}

\usepackage{microtype}

\usepackage{inconsolata}

\usepackage{tikz} 
\usepackage{threeparttable}
\usepackage{array}
\usepackage{multirow}
\usepackage{makecell}
\usepackage{amssymb}
\usepackage{changes}
\usepackage{amsmath}

\title{\textsc{FactKG}: Fact Verification via Reasoning on Knowledge Graphs}

\author{Jiho Kim$^1$, Sungjin Park$^1$, Yeonsu Kwon$^1$, Yohan Jo$^2$\thanks{\hspace{0.2cm}This work is not associated with Amazon.}, James Thorne$^1$, Edward Choi$^1$\\
  $^1$KAIST $^2$Amazon \\
  \texttt{\{jiho.kim, zxznm, yeonsu.k, thorne, edwardchoi\}@kaist.ac.kr} \\
  \texttt{jyoha@amazon.com}
  }

\newcounter{jtCounter}

\begin{document}
\maketitle

\begin{abstract}

In real world applications, knowledge graphs (KG) are widely used in various domains (e.g. medical applications and dialogue agents). However, for fact verification, KGs have not been adequately utilized as a knowledge source. 
KGs can be a valuable knowledge source in fact verification due to their reliability and broad applicability. 
A KG consists of nodes and edges which makes it clear how concepts are linked together, allowing machines to reason over chains of topics. 
However, there are many challenges in understanding how these machine-readable concepts map to information in text. 
To enable the community to better use KGs, we introduce a new dataset, \textsc{FactKG}: Fact Verification via Reasoning on Knowledge Graphs. It consists of 108k natural language claims with five types of reasoning: One-hop, Conjunction, Existence, Multi-hop, and Negation. Furthermore, \mbox{\textsc{FactKG}} contains various linguistic patterns, including colloquial style claims as well as written style claims to increase practicality.
Lastly, we develop a baseline approach and analyze \mbox{\textsc{FactKG}} over these reasoning types.
We believe \mbox{\textsc{FactKG}} can advance both reliability and practicality in KG-based fact verification.\footnote{Data available at \url{https://github.com/jiho283/FactKG}.}
\end{abstract}

\section{Introduction}

\begin{table*}[]
\resizebox{\textwidth}{!}{%
\begin{tabular}{m{3cm}|m{10cm}| >{\centering\arraybackslash}m{4cm}}
\Xhline{1.5pt}

\textbf{Reasoning Type} & \multicolumn{1}{c|}{\textbf{Claim Example}} & \textbf{Graph} \\ \Xhline{1pt}

\textbf{One-hop} & AIDAstella was built by Meyer Werft. &
\begin{tikzpicture}[node distance={15mm}, main/.style = {draw, circle}] 
\node[main] (1){$s$};
\node[main] (2) [right of=1] {$m$}; 
\draw[->] (1) -- node[midway, above] {$r_2$} (2) ;
\end{tikzpicture}  \\ \hline

\textbf{Conjunction} &  AIDA Cruise line operated the AIDAstella which was built by Meyer Werft.&
\begin{tikzpicture}[node distance={15mm}, main/.style = {draw, circle}] 
\node[main] (1){$c$};
\node[main] (2) [right of=1] {$s$}; 
\node[main] (3) [right of=2] {$m$};
\draw[->] (2) -- node[midway, above] {$r_3$} (1) ;
\draw[->] (2) -- node[midway, above] {$r_2$} (3);
\end{tikzpicture}  \\ \hline

\textbf{Existence} &  Meyer Werft had a parent company. &
\begin{tikzpicture}[node distance={15mm}, main/.style = {draw, circle}] 
\node[main] (1){$m$};
\node[main, white] (2) [right of=1] {}; 
\draw[->, dashed] (1) -- node[midway, above] {$r_1$} (2);
\end{tikzpicture}  \\ \hline

\textbf{Multi-hop} &  AIDAstella was built by a company in Papenburg.& 
\begin{tikzpicture}[node distance={15mm}, main/.style = {draw, circle}] 
\node[main] (1){$s$};
\node[main, dashed] (2) [right of=1] {$x$}; 
\node[main] (3) [right of=2] {$p$};
\draw[->] (1) -- node[midway, above] {$r_2$} (2);
\draw[->] (2) -- node[midway, above] {$r_4$} (3);
\end{tikzpicture}  \\ \hline

\textbf{Negation} &  AIDAstella was \textcolor{red}{not} built by Meyer Werft in Papenburg.&
\begin{tikzpicture}[node distance={15mm}, main/.style = {draw, circle}] 
\node[main] (1){$s$};
\node[main] (2) [right of=1] {$m$}; 
\node[main] (3) [right of=2] {$p$};
\draw[->, red] (1) -- node[midway, above] {$r_2$} (2);
\draw[->] (2) -- node[midway, above] {$r_4$} (3);
\end{tikzpicture}  \\ \Xhline{1.5pt}
\end{tabular}%
}
\caption{Five different reasoning types of \textsc{FactKG}. $r_1$: parentCompany, $r_2$: shipBuilder, $r_3$: shipOperator, $r_4$: location, $m$: Meyer Werft, $s$: AIDAstella, $c$: AIDA Cruises. 
\label{tab:types}
}
\vspace{-1mm}
\end{table*}

\begin{figure}[t]
    \begin{center}
    \includegraphics[width=1.0\linewidth]{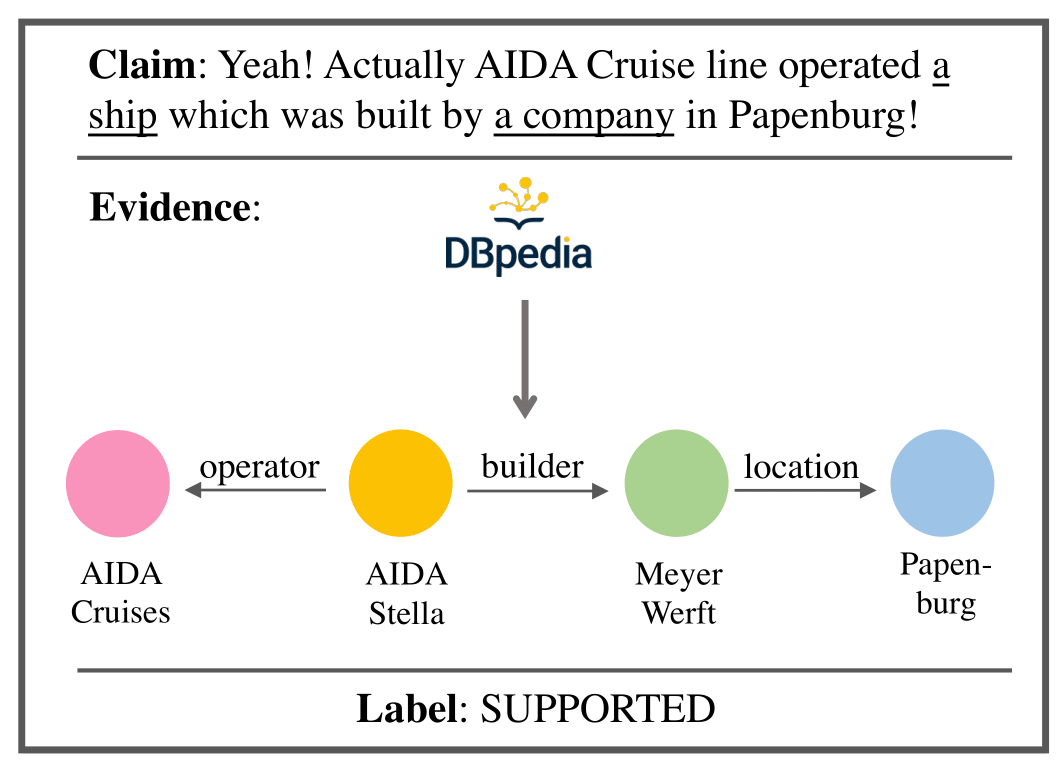}
    \end{center}
    \caption{An example data from \textsc{FactKG}. To verify the claim whether it is \textsc{Supported} or \textsc{Refuted}, we use triples extracted from DBpedia as evidence.}
    \label{figure1}
    \vspace{-3mm}
\end{figure}

The wide spread risk of misinformation has increased the demand for fact-checking, that is, judging whether a claim is true or false based on evidence.
Accordingly, recent works on fact verification have been developed with various sources of evidence, such as text~\citep{thorne-etal-2018-fever, augenstein2019multifc, jiang2020hover, schuster-etal-2021-get, park2021faviq} and tables~\citep{chen2019tabfact, wang2021semeval, aly2021feverous}. Unfortunately, knowledge graphs (KG), one of the large-scale data forms, have not yet been fully utilized as a source of evidence. 
A KG is a valuable knowledge source due to two advantages.

Firstly, KG-based fact verification can provide more reliable reasoning: since the efficacy of real-world fact-checking hinges on this reliability, recent studies have focused on justifying the decisions of a fact verification system~\citep{kotonya-toni-2020-explainable}. 
In most existing works, the justification is based on the extractive summary of text evidence. Therefore, the inferential links between the evidence and the verdict are not clear~\citep{kotonya-toni-2020-explainable-automated, atanasova-etal-2020-diagnostic, atanasova-etal-2020-generating-fact}. 
Compared to text and tables, a KG can simply represent reasoning process with logic rules on nodes and edges~\citep{liang2022kgsurvey}.
This allows us to categorize common types of reasoning with the graphical structure, as shown in Table~\ref{tab:types}.

Secondly, KG-based fact verification techniques have broad applicability beyond the domain of fact-checking. For example, modern dialogue systems (e.g. \textit{Amazon Alexa}~\citep{alexa}, \textit{Google Assistant}~\citep{kale2018knowledge}) maintain and communicate with internal knowledge graphs,
and it is crucial to make sure that their content is consistent with what the user says and otherwise update the knowledge graphs accordingly. If we model the user's utterance as a claim and the dialogue system's internal knowledge graph as a knowledge source, the process of checking their consistency can be seen as a form of KG-based fact verification task.
More generally, KG-based fact verification techniques can be applied to cases which require checking the consistency between graphs and text.

Reflecting these advantages, we introduce a new dataset, \textsc{FactKG}: Fact Verification via Reasoning on Knowledge Graphs, consisting of 108k textual claims that can be verified against DBpedia~\citep{lehmann2015dbpedia} and labeled as \textsc{Supported} or \textsc{Refuted}. We generated the claims based on graph-text pairs from WebNLG~\cite{gardent2017webnlg} to incorporate various reasoning types.
The claims in \textsc{FactKG} are categorized into five reasoning types: One-hop, Conjunction, Existence, Multi-hop, and Negation. Furthermore, \textsc{FactKG} consists of claims in various styles including colloquial, making it potentially suitable for a wider range of applications, including dialogue systems.

We conducted experiments on \textsc{FactKG} to validate whether graph evidence had a positive effect for fact verification.
Our experiments indicate that the use of graphical evidence in our model resulted in superior performance when compared to baselines that did not incorporate such evidence.

\section{Related Works}
\begin{figure*}[t]
    \begin{center}
    \includegraphics[width=1.0\linewidth]{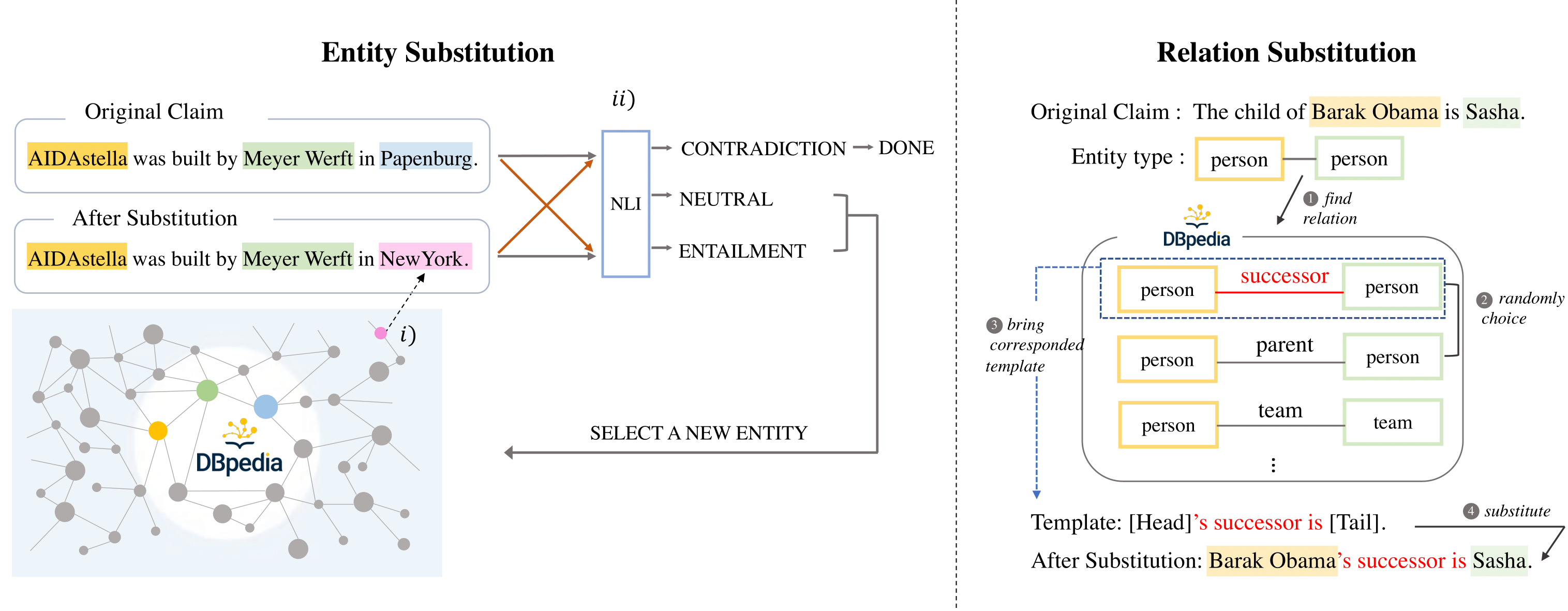}
    \end{center}
    \caption{Two substitution methods utilized in \textsc{FactKG}. In \textit{Entity substitution}, we select a new entity located in outside 4-hops from all entities in the original claim. If the results of bidirectional NLI are both contradiction, we finish this process. In \textit{Relation substitution}, we randomly extract a relation that takes the same entity types for the head and tail as the original relation. Then, substitution is performed based on a template specific to the selected relation.}
    \label{figure2}
\end{figure*}

\subsection{Fact Verification and Structured Data}

There are various types of knowledge used in fact verification such as text, tables, and knowledge graphs. 
Research on fact verification has mainly focused on text data as evidence~\citep{thorne-etal-2018-fever, augenstein2019multifc, jiang2020hover, schuster-etal-2021-get, park2021faviq}. FEVER~\citep{thorne-etal-2018-fever}, one of the representative fact verification datasets, is a large-scale manually annotated dataset derived from Wikipedia. 
Other recent works leverage ambiguous QA pairs~\citep{park2021faviq}, factual changes~\citep{schuster-etal-2021-get}, multiple documents~\cite{jiang2020hover}, or claims sourced from fact checking websites~\citep{augenstein2019multifc}. 
Fact verification on table data is also studied~\citep{chen2019tabfact, wang2021semeval, aly2021feverous}. Table-based datasets such as SEM-TAB-FACTS~\citep{wang2021semeval} or TabFact~\citep{chen2019tabfact} require reasoning abilities over tables, and FEVEROUS~\citep{aly2021feverous} validate claims utilizing table and text sources. We refer the reader to \citet{guo-etal-2022-survey} for a comprehensive survey.

There have been several tasks that utilize knowledge graphs~\citep{dettmers2018convolutional}. For example, FB15K~\citep{NIPS2013_1cecc7a7}, FB15K-237~\citep{2015-observed}, and WN18~\citep{NIPS2013_1cecc7a7} are built upon subsets of large-scale knowledge graphs, Freebase~\citep{bollacker2008freebase} and WordNet~\citep{miller1995wordnet} respectively. These datasets only use a single triple as a claim, and thus the claims only require One-hop reasoning. However, \textsc{FactKG} is the first KG-based fact verification dataset with natural language claims that require complex reasoning. In terms of the evidence KG size, \textsc{FactKG} uses the entire DBpedia (0.1B triples), which is significantly larger than previous datasets (FB15K: 592K, FB15K-237: 310K, WN18: 150K).

\subsection{WebNLG}
As constructing a KG-based fact verification dataset requires a paired text-graph corpus, we utilized WebNLG as a basis for \textsc{FactKG}. 
WebNLG is a dataset for evaluating triple-based natural language generation, which consists of 25,298 pairs of high-quality text and RDF triples from DBpedia.
WebNLG contains diverse forms of graphs and the texts are created by linguistic experts, which gives it great variety and sophistication. 
In the 2020 challenge\footnote{\url{https://webnlg-challenge.loria.fr/challenge_2020/}}, the dataset has been expanded to 45,040 text-triples pairs. We used this 2020 version of WebNLG when constructing our dataset.

\section{Data Construction}

Our goal is to diversify the graph reasoning patterns and linguistic styles of the claims. To achieve this, we categorize five reasoning types of claims: One-hop, Conjunction, Existence, Multi-hop, and Negation.
Our claims are generated by transforming the sentences in $S_w$, a subset of WebNLG's text-graph pairs (Section \ref{sec:gen}).\footnote{We found that 99.7\% of claims in FEVER and FEVEROUS consist of a single sentence. 
To reflect this result, we extract a subset $S_w$ containing only single sentences from WebNLG.} 
Next, we also diversified the claims with colloquial style transfer and presupposition (Section \ref{sec:coll}).

\subsection{Claim Generation}
\label{sec:gen}

\subsubsection{One-hop}
\label{sec:onehop}

The most basic type of claim is one-hop, which covers only one knowledge triple. 
One-hop claims can be verified by checking the existence of a single corresponding triple. In the second row of Table~\ref{tab:types}, the claim is \textsc{Supported} when the triple (\textit{AIDAstella}, \textit{ShipBuilder}, \textit{Meyer Werft}) exists.

We take the sentences that consist of a single triple in $S_w$ as \textsc{Supported} claims. \textsc{Refuted} claims are created by substituting \textsc{Supported} claims in two ways: \textit{Entity substitution} and \textit{Relation substitution}. 
In \textit{Entity substitution}, we replace an entity $e$ in \textsc{Supported} claim $C$ with another entity $\tilde{e}$ of the same entity type. 
In order to ensure that the label of the substituted sentence $\tilde{C}$ is \textsc{Refuted}, the entity $\tilde{e}$ should satisfy the following two conditions.
\textit{i)} To select $\tilde{e}$ that is irrelevant to $C$, $\tilde{e}$ is outside 4-hops from all entities in $C$ on DBpedia, \textit{ii)} the results of NLI ($C$, $\tilde{C}$) and NLI ($\tilde{C}$, $C$) are both \textsc{contradiction}.\footnote{We use a natural language inference (NLI) model, RoBERTa-base~\citep{DBLP:journals/corr/abs-1907-11692} finetuned on the MNLI dataset~\citep{N18-1101}. The notation NLI (\textit{p}, \textit{h}) represents the result of NLI when \textit{p} is assigned as the premise and \textit{h} as the hypothesis.} 
In \textit{Relation substitution}, we replace a relation in the \textsc{Supported} claim with another relation. 
We replace the relation of a triple in the claim with another relation that takes the same entity types for the head and tail as the original relation (e.g. currentTeam $\leftrightarrow$ formerTeam). The four groups of compatible relations are listed in Table \ref{tab:apprelsub}.
The overall process of the substitution methods is illustrated in Figure~\ref{figure2}.

\subsubsection{Conjunction}
\label{sec:conj}
A claim in the real world can include a mixture of different facts. To incorporate this, we construct a conjunction claim composed of multiple triples. Conjunction claims are verified by the existence of all corresponding triples. In the third row of Table~\ref{tab:types}, the claim can be divided into two parts: \textit{``AIDA Cruise line operated the AIDAstella.''} and \textit{``AIDAstella was built by Meyer Werft.''}. The claim is \textsc{Supported} when all the triples (\textit{AIDAstella}, \textit{ShipOperator}, \textit{AIDA Cruises}), (\textit{AIDAstella}, \textit{ShipBuilder}, \textit{Meyer Werft}) exist.
To implement this idea, we extracted sentences consisting of more than one triple from $S_w$ and used them as the \textsc{Supported} claims.
To create \textsc{Refuted} claims, we use \textit{Entity substitution} method on these \textsc{Supported} claims.

\begin{figure}[t]
    \begin{center}
    \includegraphics[width=\linewidth]{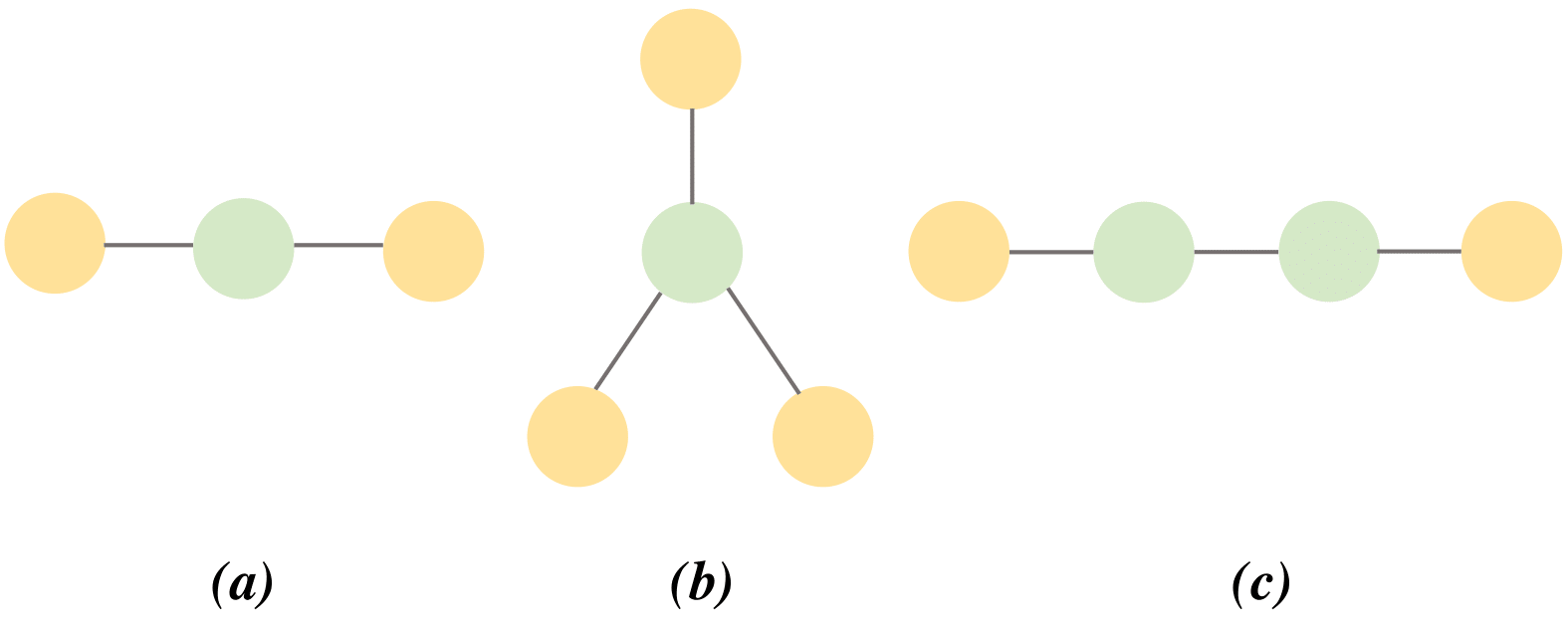}
    \end{center}
    \caption{Graph patterns used in Conjunction and Multi hop claims.}
    \label{figure3}
\end{figure}

\subsubsection{Existence}
\label{sec:exist}
People may make claims that assert the existence of something (e.g. \textit{``She has two kids.''}).
From the view of a triple, this corresponds to the head or tail missing.
To reflect this scenario, we formulate a claim by extracting only \{\textit{head}, \textit{relation}\} or \{\textit{tail}, \textit{relation}\} from a triple. 
Existence claims are generated using templates and they are divided into two categories: \textit{head-relation} (e.g. template: \{\textit{head}\} had a(an) \{\textit{relation}\}.) and \textit{tail-relation} (e.g. template: \{\textit{tail}\} was a \{\textit{relation}\}.). \textsc{Supported} claims are constructed by randomly extracting \{\textit{head}, \textit{relation}\} or \{\textit{tail}, \textit{relation}\} in triples from $S_w$. 
The \textsc{Refuted} claims are constructed using the same type of entities as represented in the claim, but with different relations.
However, it is possible that unrealistic claims may be generated in this manner. For example, \textit{``Meyer Werft had a location.''} or \textit{``Papenburg was a location.''} can be created from the triple (\textit{Meyer Werft}, \textit{location}, \textit{Papenbug}). Hence, we selected 22 relations out of all relations that lead to realistic claims.
Templates used for both categories and examples of generated claims are in Table~\ref{tab:apprel}.

\subsubsection{Multi-hop}
\label{sec:multihop}
We also consider multi-hop claims that require the validation of multiple facts where some entities are underspecified. 
Entities in this claim can be connected by a sequence of relations.
For example, the multi-hop claim in Table~\ref{tab:types} is \textsc{Supported} if the triple (\textit{AIDAstella}, \textit{ShipBuilder}, $x$) and the triple ($x$, \textit{location}, \textit{Papenburg}) are present in the graph. The goal is to verify the existence of a path on the graph that starts from \textit{AIDAstella} and reaches \textit{Papenburg} through the relations \textit{ShipBuilder} and \textit{location}.

Figure~\ref{figure3} shows how a \textsc{Supported} multi-hop claim $C_M$ can be generated by replacing an entity $e$ of the conjunction claim $C$ with its type name. 
First, an entity $e$ is selected from the green nodes. Then, the type name $t$ of the entity $e$ is extracted from DBpedia. 
However, each entity $e$ in DBpedia has several types $T = \{t_1,t_2,...,t_N\}$, and it is not annotated which type is relevant when $e$ is used in a claim. So it is necessary to select one of them. 
For each $t_n \in T$, we insert it next to the entity $e$ in the claim $C$ and measure the perplexity score of the modified claim using GPT2-large~\citep{radford2019language}.
Then we replace $e$ in the claim with the type name that had the lowest score.
The \textsc{Refuted} claim is generated by applying \textit{Entity substitution} to the \textsc{Supported} claim.

\subsubsection{Negation}
\label{sec:neg}
For each of the four methods for generating claims, we develop claims that incorporate negations.

\textbf{One-hop}\hspace{5mm}
We use the Negative Claim Generation Model~\citep{lee2021crossaug} which was fine-tuned on the opposite claim set in the WikiFactCheck-English dataset~\citep{wikifactchkeng:2020:LREC}.\footnote{WikiFactCheck-English consists of pairs of claims, with a positive claim and its corresponding negative claim.} To ensure the quality of the generated sentences, we generate 100 opposing claims for each original claim, then only use those that preserve all entities, and contain negations (e.g. \textit{`not'} or \textit{`never'}).
Also, similar to \textit{Entity substitution} method, we only use sentences whose NLI relation with the original sentences are \textsc{contradiction} bidirectionally.
When a negation is added, the label of the generated claim is reversed from the original claim.

\textbf{Conjunction}\hspace{5mm}
The use of negations (i.e., \textit{`not'}) in various positions within conjunction claims allows the generation of a wide range of negative claim structures.
We employ the pretrained language model GPT-J 6B~\citep{gpt-j} to attach negations to the claim. 
We construct 16 in-context examples, each with negations attached to the texts corresponding to the first or/and second relation. 
When a negation is added to the \textsc{Supported} claims, all the claims become \textsc{Refuted}.
However, when it is added to \textsc{Refuted} claims, the label depends on the position of the negation. 
When negations are added to all parts with substituted entities, it becomes a \textsc{Supported} claim.
Conversely, other cases preserve the label \textsc{Refuted} since the negation is added to a place that is not related to entity substitution.
A detailed labeling strategy is described in Appendix~\ref{app:negconj}.

\textbf{Existence}\hspace{5mm}
The claim is formulated by adding a negation within the templates presented in Section \ref{sec:exist} (e.g. \{\textit{tail}\} was not a \{\textit{relation}\}.).

\textbf{Multi-hop}\hspace{5mm}
A claim is formulated using the GPT-J with in-context examples, similar to conjunction.
The truth of this claim is dependent on the presence of a distinctive path that matches the claim's intent.
For example, the negative claim \textit{``AIDAstella was built by a company, not in Papenburg.''} is \textsc{Supported} if $x$ exists where the triples (\textit{AIDAstella}, \textit{ShipBuilder}, $x$) and ($x$, \textit{location}, $y$) are in DBpedia and $y$ is not \textit{Papenburg}. A more detailed labeling strategy is in Appendix~\ref{app:negmul}.

\subsection{Colloquial Style Transfer}
\label{sec:coll}
We transform the claims into a colloquial style via style transfer using both a fine-tuned language model and presupposition templates.

\subsubsection{Model based}
Using a similar method proposed by \citet{2021-robust}, we transform the claim obtained from \ref{sec:gen} into a colloquial style. For example, the claim \textit{``Obama was president.''} is converted to \textit{``Have you heard about Obama? He was president!''}.

We train FLAN T5-large~\cite{https://doi.org/10.48550/arxiv.2210.11416} to generate a colloquial style sentence given a corresponding written style sentence from Wizard of Wikipedia~\cite{dinan2019wizard}.
However, using sentences generated by the model could have several potential issues: \textit{i)} the original and generated sentences are lexically the same, \textit{ii)} some entities are missing in the generated sentences, \textit{iii)}  the generated sentences deviate semantically from the original, \textit{iv)} the generated sentences lack a colloquialism, as mentioned in 
 \citet{2021-robust}. To overcome this, we oversample candidate sentences and utilize an additional filtering process.

First, to make more diverse samples using the model, we set the temperature to 20.0 and generate 500 samples with beam search. \textit{i)} To avoid generated sentences that are too similar to the original sentences, only sentences with an edit distance of 6 or more from the original sentence are selected among 500 samples.
\textit{ii)} Then, only those that have verbs and the named entities all preserved are selected.\footnote{NLTK~\cite{bird2009natural} POS tagger and Stanza~\cite{qi2020stanza} NER module are used. DBPedia entities are already tagged in each claim, but not all entities exist in the sentence in their raw form, so the NER module is used.}
\textit{iii)} Finally, we use bidirectional NLI to preserve the original semantics.
Candidate sentences survive when NLI ($O$, $G$) is \textsc{entailment} and NLI ($G$, $O$) is not \textsc{contradiction} where $O$ refers to the original sentence and $G$ the generated sentence.
On average, only 41.2 generated sentences survived out of 500 samples.
Additionally, in cases where none of the 500 generated sentences pass the filtering process, we include the original claim in the final dataset as a written style claim. Following the filtering process, the AFLITE method \cite{sakaguchi2019winogrande}, which utilizes adversarial filtering, is applied to select the most colloquial style sentence among the surviving sentences. We include the selected claim in the final dataset as a colloquial style claim.

\subsubsection{Presupposition}
A presupposition is something the speaker assumes to be the case prior to making an utterance~\cite{yule1996pragmatics}. 
People often communicate under the presupposition that their beliefs are universally accepted. 
We construct claims using this form of utterance.
The claims in \textsc{FactKG} are focused on three types of presupposition: factive, non-factive, and structural presuppositions.

\textbf{Factive Presupposition}\hspace{5mm} 
People frequently use verbs like \textit{``realize''} or \textit{``remember''} to express the truth of their assumptions.
The utterance \textit{``I remembered that }\{Statement\}\textit{.''} assumes that \{Statement\} is true. 
Reflecting these features, a new claim is created by appending expressions that contain presupposition (e.g. \textit{``I realized that'' or ``I wasn't aware that''}) to the existing claim.
We used eight templates to make factive presupposition claims: the details are appended in Table~\ref{tab:factpresup}.

\textbf{Non Factive Presupposition}\hspace{5mm} The verbs such as \textit{``wish''} are commonly used in utterances that describe events that have not occurred.
For example, people say \textit{``I wish that }\{Statement\}\textit{.''} when \{Statement\} did not happen.
Claims that are created by the non-factive presupposition method are labeled as the opposite of the original one. 
We used three templates to make these claims: the templates are appended in Table~\ref{tab:factpresup}.

\textbf{Structural Presupposition}\hspace{5mm} 
This type is in the form of a question that presumes certain facts. We treat the question itself as a claim. For example, \textit{``When was Messi in Barcelona?''} assumes that \textit{Messi was in Barcelona}. 
To create a natural sentence form, only claims corresponding to one-hop and existence are constructed. 
For the one-hop claim, a different template was created corresponding to each relation reflecting its meaning (e.g. ``When did \{\textit{head}\} die from \{\textit{tail}\}?'' for the relation \textit{deathCause} and ``When was \{\textit{head}\} directed by \{\textit{tail}\}?'' for relation \textit{director}).
Existence claims are also generated based on templates (e.g. ``When was \{\textit{tail}\} \{\textit{relation}\}?'') using pairs of \textit{head}-\textit{relation} or \textit{tail}-\textit{relation}, similar to Section \ref{sec:exist}.
The templates used are described in Table \ref{tab:strcutprep}.

\subsection{Quality Control}

To evaluate the quality of our dataset, the labeling strategy and the output of the colloquial style transfer model are assessed.

\textbf{Labeling Strategy}\hspace{5mm} When \textsc{Supported} claims are made in the manner described in Section \ref{sec:gen}, the labeling is straightforward, as all have precise evidence graphs.
However, \textsc{Refuted} claims are generated by random substitution, so there might be a small chance that they remain \textsc{Supported} (e.g. \textit{``The White House is in Washington, D.C.''} to \textit{``The White House is in America.''}).
To evaluate this substitution method, randomly sampled 1,000 substituted claims were reviewed by two graduate students.
As a result, 99.4\% of generated claims were identified as \textsc{Refuted} by both participants.

\textbf{Colloquial Style Transfer Model}\hspace{5mm} We also evaluate the quality of the colloquial style claims generated by the model.
A survey was conducted on all claims in the test set by three graduate students. 
As a result, only 9.8\% of the claims were selected as \textit{Loss of important information} by at least two reviewers.
In addition, to ensure the quality of the test set, only claims that were selected as \textit{All facts are preserved} by two or more reviewers are included in the test set. The survey details are in Appendix \ref{app:survey}. 

\label{sec:dcon}

\section{Experiments}
\begin{table}[]
\resizebox{\columnwidth}{!}{%
\begin{tabular}{ccccc}
\Xhline{1.5pt}
\multirow{2}{*}{\textbf{Type}} & \multirow{2}{*}{\textbf{Written}} & \multicolumn{2}{c}{\textbf{Colloquial }} & \multirow{2}{*}{\textbf{Total}} \\ \cline{3-4}
                               &                                   & \textbf{Model}   & \textbf{Presup}   &                                 \\ \hline
One-hop                        & 2,106                             & 15,934           & 1,580                     & 19,530                          \\ 
Conjuction                     & 20,587                            & 15,908           & 602                       & 37,097                          \\ 
Existence                      & 280                               & 4,060            & 4,832                     & 9,172                           \\ 
Multi-hop                      & 10,239                            & 16,420           & 603                       & 27,262                          \\ 
Negation                       & 1,340                             & 12,466           & 1,807                     & 15,613                          \\ 
\hline
Total                          & 34,462                            & 64,788           & 9,424                     & 108,674                         \\ \Xhline{1.5pt}
\end{tabular}
}
\caption{Dataset statistics of \textsc{FactKG} for all reasoning types.
\label{tab:stats}
}
\end{table}

\begin{figure*}[t]
    \begin{center}
    \includegraphics[width=\linewidth]{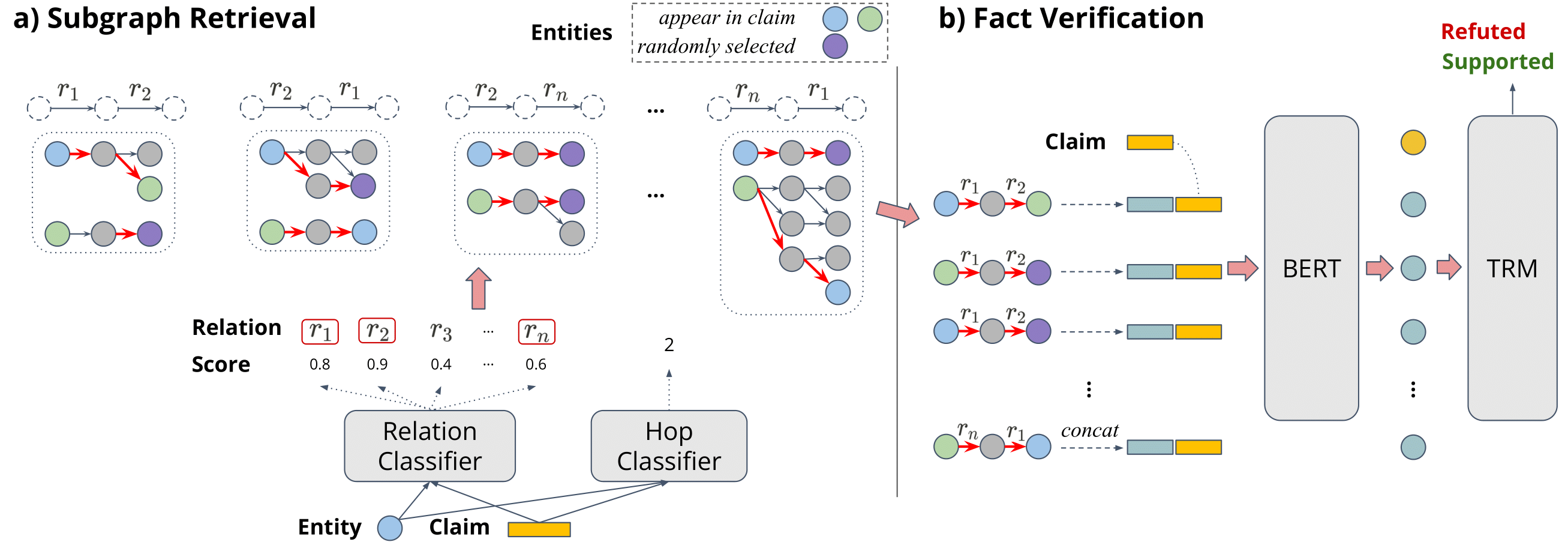}
    \end{center}
    \caption{Overall process of our baseline. In the subgraph retrieval step, each classifier respectively predicts the relations and hops related to the given entity and the claim. Subsequently, we check all the n-hop relation sequences obtained from each classifier to find all evidence paths. In the fact verification step, the claim is verified by leveraging all outputs obtained from the subgraph retrieval step. In this figure, we denote Transformer Encoder as TRM.}
    \label{figure4}
\end{figure*}

\subsection{Dataset Statistics}
\label{data:stat}
Table~\ref{tab:stats} shows the statistics of \textsc{FactKG}.
We split the claims into train, dev, and test sets with a proportion of 8:1:1.
We ensured that the set of triples in each split is disjoint with the ones in other splits.

\subsection{Experimental Setting}
We publish \textsc{FactKG} with sets of claims, graph evidence and labels. 
The graph evidence includes entities and a set of relation sequences connected to them.
For instance, when the claim is given as \textit{``AIDAstella was built by a company in Papenburg.''}, the entity \textit{`AIDAstella'} corresponds to a set of relation sequence [\textit{shipBuilder}, \textit{location}] and \textit{`Papenburg'} corresponds to [\textit{$\sim$location}, \textit{$\sim$shipBuilder}].\footnote{`$\sim$' indicates that the direction of the relation is reversed}
In the test set, we only provide entities as graph evidence.
\label{sec:expset}

\subsection{Baseline}

We conduct experiments on \textsc{FactKG} to see how the graphical evidence affects the fact verification task.
To this end, we divided our baselines into two distinct categories based on the input type, \textit{Claim Only} and \textit{With Graphical Evidence}.
\subsubsection{Claim Only}

In the \textit{Claim Only} setting, the baseline models receive only the claim as input and predict the label.
We used three transformer-based text classifiers, BERT, BlueBERT, and Flan-T5.
BERT~\citep{bertdevlin} is trained on Wikipedia from which DBpedia is extracted.
So we expect that the model will use evidence memorized in its pre-trained weights~\citep{petroni-etal-2019-language} or exploit structural patterns in the generated claims \citep{2019-towards-debiasing,thorne-vlachos-2021-elastic}.
BlueBERT~\citep{peng2019transfer} is trained on biomedical corpus, such as Pubmed abstracts. 
We use BlueBERT as a comparator for BERT since it has never seen Wikipedia during its pre-training.
Flan-T5~\citep{https://doi.org/10.48550/arxiv.2210.11416} is an enhanced version of T5~\citep{raffel2019t5} encoder-decoder that has been fine-tuned in a mixture of 1.8K tasks.
In all experiments, we fine-tune BERT and BlueBERT on our training set. 
Different from BERT and BlueBERT, we use Flan-T5 in the zero-shot setting. For this setting, we use \textit{``Is this claim True or False? Claim: ''} as the prefix. Then, we measure the probability that tokens \textit{True} and \textit{False} will appear in the output. 
Among the two tokens, we choose the one with the higher probability.

\subsubsection{With Graphical Evidence}
In the \textit{With Graphical Evidence} setting, the model receives the claim and graph evidence as input and predicts the label. 
The baseline we used is a framework proposed by GEAR~\citep{zhou-etal-2019-gear} that enables reasoning on multiple evidence texts.
Since GEAR was originally designed to reason over text passages, we change components to suit KG.
The modified GEAR consists of the subgraph retrieval module and the claim verification module.
The pipeline of the modified GEAR is illustrated in Figure~\ref{figure4}.

\noindent \textbf{Subgraph retrieval}\hspace{3mm} We replace document retrieval and sentence selection in GEAR with subgraph retrieval. 
To retrieve graphical evidence, we train two independent BERT models, namely a relation classifier and a hop classifier.
The relation classifier predicts the set of relations $R$ from the claim $c$ and the entity $e$.
The hop classifier is designed to predict the maximum number of hops $n$ to be traversed from $e$. 
We take the subgraph of $G$ that are composed only of the relations in $R$ and where the terminal nodes are entities in $C$ and less than $n$ hops apart from $e$, allowing for duplicates and considering the order.
By traversing the knowledge graph starting from $e$ along the relation sequences in $P$, we choose the paths that can reach another entity that appears in the claim.
If none of the paths is reachable to other entities, then we randomly choose one of the paths.
The strategy we used enables the model to retrieve supported evidence and counterfactual evidence for the given claim.
The following example is presented to assist the understanding of our subgraph retrieval method.
The example claim in Section~\ref{sec:expset} consists of two entities, \textit{`AIDAstella'} and \textit{`Papenburg'}.
In this setting, the hop classifier must predict 2 since those entities are connected by a sequence of two relations, namely \textit{shipBuilder} and \textit{location}.
In addition, the relation classifier must predict correctly predict those two relations.
After that, we find all 2-hop paths starting from \textit{`AIDAstella'} along the predicted relations in the knowledge graph.
If there is a path that reaches \textit{`Papenburg'}, we can use it as supporting evidence. 
If not, however, we randomly select a path. 

\begin{table*}[]
\centering
\resizebox{0.9\textwidth}{!}{%
\begin{tabular}{c|c|ccccc|c}
\Xhline{1.5pt} 
\textbf{Input Type}                  & \textbf{Model} & \textbf{One-hop} & \textbf{Conjunction} & \textbf{Existence} & \textbf{Multi-hop} & \textbf{Negation} & \textbf{Total} \\ \hline
\multirow{3}{*}{\textit{Claim Only}} & BERT           & 69.64            & 63.31               & 61.84              & 70.06              & 63.62             & 65.20          \\
                                     & BlueBERT       & 60.03            & 60.15               & 59.89              & 57.79              & 58.90             & 59.93          \\
                                     & Flan-T5        & 62.17            & 69.66               & 55.29              & 60.67              & 55.02             & 62.70          \\ \hline
\textit{With Evidence}               & GEAR           & 83.23            & 77.68               & 81.61              & 68.84              & 79.41             & 77.65          \\ \Xhline{1.5pt} 
\end{tabular}
}
\caption{Fact verification accuracy on \textsc{FactKG}.
\label{tab:results}
}
\end{table*}

\par \noindent \textbf{Fact verification}\hspace{3mm}
We directly employed the claim verification in GEAR and applied some changes to suit the KG setting. Since our evidence is a set of graph paths, we converted them to text by concatenating each triple with the special token <SEP>.
We also found that ERNet in GEAR is identical to the Transformer encoder, so we replaced it with a randomly initialized Transformer encoder. 
To make this paper self-contained, we provide further details about the claim verification of GEAR in Appendix~\ref{app:gear}.

\subsection{Results}
\paragraph{Fact Verification Results}
We evaluated the performance of the models in predicting labels and reported the accuracy in Table~\ref{tab:results} by different reasoning types. 

As we expected, GEAR outperforms other baseline models in most of reasoning types because it used graph evidence. Especially, in existence and negation, GEAR substantially outperforms \textit{Claim Only} baselines. 
Since the existence claims contain significantly less information than other types, having to search for evidence seems to increase fact verification performance. 
In addition, negation claims require additional inference steps compared to other types, thus logical reasoning based on graph evidence would help the model make correct prediction.

In the multi-hop setting, however, the accuracy of GEAR is lower than BERT, which may be due to the increased complexity of graph retrieval.
When entities are far apart with many intermediate nodes being under-specified, it increases the probability of retrieving an incorrect graph.
In GEAR, text and evidence paths are concatenated and used as input, so if many incorrect graphs are retrieved, they can lead to incorrect predictions. 
Also, the accuracy of BERT is the most superior in the multi-hop setting, which suggests that masked language modeling facilitates the model to robustly handle unspecified entities in the multi-hop claims.

In the \textit{Claim Only} setting, all baselines outperform the Majority Class (51.35\%), and the BERT model shows the highest performance.
BlueBERT was pre-trained in the same manner, but BERT shows superior performance due to its pre-trained knowledge from Wikipedia.

\begin{table}[]
\centering
\resizebox{\linewidth}{!}{%
\begin{tabular}{c|c|cc|cc}
\Xhline{1.5pt} 
\textbf{Input Type}                  & \textbf{Model} & \textbf{W $\rightarrow$ W} & \textbf{W $\rightarrow$ C} & \textbf{C $\rightarrow$ C} & \textbf{C $\rightarrow$ W} \\ \hline
\multirow{2}{*}{\textit{Claim Only}} & BERT         &  71.75        & 63.85         &  68.10         & 69.43       \\
                                     & BlueBERT     &  64.76        & 56.28         &  58.77        & 63.92       \\ \hline
\textit{With Evidence}               & GEAR         &  81.00        & 75.43         &  80.81        & 78.80       \\ \Xhline{1.5pt} 
\end{tabular}
}
\caption{\textbf{W} refers to written style claims and \textbf{C} refers to colloquial style claims. \textbf{W $\rightarrow$ C} means that the model is trained on the written style claim set and tested on the colloquial style claim set. Flan-T5 is not used in this experiment because we use it only in the zero-shot setting.
\label{tab:resultrans}
}
\end{table}

\paragraph{Cross-Style Evaluation}
We split the dataset into two disjoint sets, written style and colloquial style.
We perform a cross-style fact verification task by using those datasets and the results are reported in Table~\ref{tab:resultrans}.

Initially, we anticipated that using different styles for the train and test set would result in a significant decrease in verification performance. 
However, contradict our expectation, in \textbf{C$\rightarrow$W} setting, BERT and BlueBERT show an improvement in performance over \textbf{C$\rightarrow$C}.
Even in GEAR, the performance score only dropped slightly.
Therefore, the results demonstrate that colloquial style is constructed in various forms which can be beneficial for generalization.

\label{sec:exp}

\section{Conclusion}
In this paper, we present \textsc{FactKG}, a new dataset for fact verification using knowledge graph.
In order to reveal the relationship between fact verification and knowledge graph reasoning, we generated claims corresponding to a certain graph pattern.
Additionally, \textsc{FactKG} also includes colloquial-style claims that are applicable to the dialogue system.
Our analysis showed that the claims in our dataset are difficult to solve without reasoning over the knowledge graph.

We expect the dataset to offer various research directions. 
One possible use of our dataset is as a benchmark for justification prediction.
Most research on this task generate a text passage as justification, yet this approach lacked a gold reference.
On the contrary, the interpretability of the knowledge graph allows us to employ it as an explanation for the verdict, such as question answering in the medical domain where explainability is important. Furthermore, using the KG structure for the claim generation allows us to generate a dataset with more complex multi-hop reasoning by design without relying on annotator creativity.

\label{sec:conclu}

\section*{Limitations}
Since WebNLG is derived from 2015-10 version of DBpedia, \textsc{FactKG} does not reflect the latest knowledge. Also, another limitation of our work is that the claims of \textsc{FactKG} are constructed based on single sentences, like other crowdsourced fact verification datasets. If the claim is generated by more than one sentences, the dataset will be more challenging.
We remain this challenging point as a future work.

\section*{Acknowledgements}
This work was supported by the Institute of Information \& Communications Technology Planning \& Evaluation (IITP) grant (No.2019-0-00075, No.2022-0-00984), and National Research Foundation of Korea (NRF) grant (NRF-2020H1D3A2A03100945), funded by the Korea government (MSIT).

\bibliography{anthology,custom}

\begin{thebibliography}{43}
\expandafter\ifx\csname natexlab\endcsname\relax\def\natexlab#1{#1}\fi

\bibitem[{Aly et~al.(2021)Aly, Guo, Schlichtkrull, Thorne, Vlachos,
  Christodoulopoulos, Cocarascu, and Mittal}]{aly2021feverous}
Rami Aly, Zhijiang Guo, Michael Schlichtkrull, James Thorne, Andreas Vlachos,
  Christos Christodoulopoulos, Oana Cocarascu, and Arpit Mittal. 2021.
\newblock Feverous: Fact extraction and verification over unstructured and
  structured information.
\newblock \emph{arXiv preprint arXiv:2106.05707}.

\bibitem[{{Amazon Staff}(2018)}]{alexa}
{Amazon Staff}. 2018.
\newblock \href
  {https://www.aboutamazon.com/news/devices/how-alexa-keeps-getting-smarter}
  {How alexa keeps getting smarter}.

\bibitem[{Atanasova et~al.(2020{\natexlab{a}})Atanasova, Simonsen, Lioma, and
  Augenstein}]{atanasova-etal-2020-diagnostic}
Pepa Atanasova, Jakob~Grue Simonsen, Christina Lioma, and Isabelle Augenstein.
  2020{\natexlab{a}}.
\newblock \href {https://doi.org/10.18653/v1/2020.emnlp-main.263} {A diagnostic
  study of explainability techniques for text classification}.
\newblock In \emph{Proceedings of the 2020 Conference on Empirical Methods in
  Natural Language Processing (EMNLP)}, pages 3256--3274, Online. Association
  for Computational Linguistics.

\bibitem[{Atanasova et~al.(2020{\natexlab{b}})Atanasova, Simonsen, Lioma, and
  Augenstein}]{atanasova-etal-2020-generating-fact}
Pepa Atanasova, Jakob~Grue Simonsen, Christina Lioma, and Isabelle Augenstein.
  2020{\natexlab{b}}.
\newblock \href {https://doi.org/10.18653/v1/2020.acl-main.656} {Generating
  fact checking explanations}.
\newblock In \emph{Proceedings of the 58th Annual Meeting of the Association
  for Computational Linguistics}, pages 7352--7364, Online. Association for
  Computational Linguistics.

\bibitem[{Augenstein et~al.(2019)Augenstein, Lioma, Wang, Lima, Hansen, Hansen,
  and Simonsen}]{augenstein2019multifc}
Isabelle Augenstein, Christina Lioma, Dongsheng Wang, Lucas~Chaves Lima, Casper
  Hansen, Christian Hansen, and Jakob~Grue Simonsen. 2019.
\newblock Multifc: A real-world multi-domain dataset for evidence-based fact
  checking of claims.
\newblock \emph{arXiv preprint arXiv:1909.03242}.

\bibitem[{Bird et~al.(2009)Bird, Klein, and Loper}]{bird2009natural}
Steven Bird, Ewan Klein, and Edward Loper. 2009.
\newblock \emph{Natural language processing with Python: analyzing text with
  the natural language toolkit}.
\newblock " O'Reilly Media, Inc.".

\bibitem[{Bollacker et~al.(2008)Bollacker, Evans, Paritosh, Sturge, and
  Taylor}]{bollacker2008freebase}
Kurt Bollacker, Colin Evans, Praveen Paritosh, Tim Sturge, and Jamie Taylor.
  2008.
\newblock Freebase: a collaboratively created graph database for structuring
  human knowledge.
\newblock In \emph{Proceedings of the 2008 ACM SIGMOD international conference
  on Management of data}, pages 1247--1250.

\bibitem[{Bordes et~al.(2013)Bordes, Usunier, Garcia-Duran, Weston, and
  Yakhnenko}]{NIPS2013_1cecc7a7}
Antoine Bordes, Nicolas Usunier, Alberto Garcia-Duran, Jason Weston, and Oksana
  Yakhnenko. 2013.
\newblock \href
  {https://proceedings.neurips.cc/paper_files/paper/2013/file/1cecc7a77928ca8133fa24680a88d2f9-Paper.pdf}
  {Translating embeddings for modeling multi-relational data}.
\newblock In \emph{Advances in Neural Information Processing Systems},
  volume~26. Curran Associates, Inc.

\bibitem[{Chen et~al.(2019)Chen, Wang, Chen, Zhang, Wang, Li, Zhou, and
  Wang}]{chen2019tabfact}
Wenhu Chen, Hongmin Wang, Jianshu Chen, Yunkai Zhang, Hong Wang, Shiyang Li,
  Xiyou Zhou, and William~Yang Wang. 2019.
\newblock Tabfact: A large-scale dataset for table-based fact verification.
\newblock \emph{arXiv preprint arXiv:1909.02164}.

\bibitem[{Chung et~al.(2022)Chung, Hou, Longpre, Zoph, Tay, Fedus, Li, Wang,
  Dehghani, Brahma, Webson, Gu, Dai, Suzgun, Chen, Chowdhery, Narang, Mishra,
  Yu, Zhao, Huang, Dai, Yu, Petrov, Chi, Dean, Devlin, Roberts, Zhou, Le, and
  Wei}]{https://doi.org/10.48550/arxiv.2210.11416}
Hyung~Won Chung, Le~Hou, Shayne Longpre, Barret Zoph, Yi~Tay, William Fedus,
  Eric Li, Xuezhi Wang, Mostafa Dehghani, Siddhartha Brahma, Albert Webson,
  Shixiang~Shane Gu, Zhuyun Dai, Mirac Suzgun, Xinyun Chen, Aakanksha
  Chowdhery, Sharan Narang, Gaurav Mishra, Adams Yu, Vincent Zhao, Yanping
  Huang, Andrew Dai, Hongkun Yu, Slav Petrov, Ed~H. Chi, Jeff Dean, Jacob
  Devlin, Adam Roberts, Denny Zhou, Quoc~V. Le, and Jason Wei. 2022.
\newblock \href {https://doi.org/10.48550/ARXIV.2210.11416} {Scaling
  instruction-finetuned language models}.

\bibitem[{Dettmers et~al.(2018)Dettmers, Minervini, Stenetorp, and
  Riedel}]{dettmers2018convolutional}
Tim Dettmers, Pasquale Minervini, Pontus Stenetorp, and Sebastian Riedel. 2018.
\newblock Convolutional 2d knowledge graph embeddings.
\newblock In \emph{Proceedings of the AAAI conference on artificial
  intelligence}, volume~32.

\bibitem[{Devlin et~al.(2018)Devlin, Chang, Lee, and Toutanova}]{bertdevlin}
Jacob Devlin, Ming{-}Wei Chang, Kenton Lee, and Kristina Toutanova. 2018.
\newblock \href {http://arxiv.org/abs/1810.04805} {{BERT:} pre-training of deep
  bidirectional transformers for language understanding}.
\newblock \emph{CoRR}, abs/1810.04805.

\bibitem[{Dinan et~al.(2019)Dinan, Roller, Shuster, Fan, Auli, and
  Weston}]{dinan2019wizard}
Emily Dinan, Stephen Roller, Kurt Shuster, Angela Fan, Michael Auli, and Jason
  Weston. 2019.
\newblock {W}izard of {W}ikipedia: Knowledge-powered conversational agents.
\newblock In \emph{Proceedings of the International Conference on Learning
  Representations (ICLR)}.

\bibitem[{Gardent et~al.(2017)Gardent, Shimorina, Narayan, and
  Perez-Beltrachini}]{gardent2017webnlg}
Claire Gardent, Anastasia Shimorina, Shashi Narayan, and Laura
  Perez-Beltrachini. 2017.
\newblock The webnlg challenge: Generating text from rdf data.
\newblock In \emph{Proceedings of the 10th International Conference on Natural
  Language Generation}, pages 124--133.

\bibitem[{Guo et~al.(2022)Guo, Schlichtkrull, and
  Vlachos}]{guo-etal-2022-survey}
Zhijiang Guo, Michael Schlichtkrull, and Andreas Vlachos. 2022.
\newblock \href {https://doi.org/10.1162/tacl_a_00454} {A survey on automated
  fact-checking}.
\newblock \emph{Transactions of the Association for Computational Linguistics},
  10:178--206.

\bibitem[{Jiang et~al.(2020)Jiang, Bordia, Zhong, Dognin, Singh, and
  Bansal}]{jiang2020hover}
Yichen Jiang, Shikha Bordia, Zheng Zhong, Charles Dognin, Maneesh Singh, and
  Mohit Bansal. 2020.
\newblock Hover: A dataset for many-hop fact extraction and claim verification.
\newblock \emph{arXiv preprint arXiv:2011.03088}.

\bibitem[{Kale and Hewavitharana(2018)}]{kale2018knowledge}
Ajinkya~Gorakhnath Kale and Sanjika Hewavitharana. 2018.
\newblock Knowledge graph construction for intelligent online personal
  assistant.
\newblock US Patent App. 15/238,679.

\bibitem[{Kim et~al.(2021)Kim, Kim, Hong, and Kim}]{2021-robust}
Byeongchang Kim, Hyunwoo Kim, Seokhee Hong, and Gunhee Kim. 2021.
\newblock \href {https://doi.org/10.18653/v1/2021.naacl-main.121} {How robust
  are fact checking systems on colloquial claims?}
\newblock In \emph{Proceedings of the 2021 Conference of the North American
  Chapter of the Association for Computational Linguistics: Human Language
  Technologies}, pages 1535--1548, Online. Association for Computational
  Linguistics.

\bibitem[{Kotonya and Toni(2020{\natexlab{a}})}]{kotonya-toni-2020-explainable}
Neema Kotonya and Francesca Toni. 2020{\natexlab{a}}.
\newblock \href {https://doi.org/10.18653/v1/2020.coling-main.474} {Explainable
  automated fact-checking: A survey}.
\newblock In \emph{Proceedings of the 28th International Conference on
  Computational Linguistics}, pages 5430--5443, Barcelona, Spain (Online).
  International Committee on Computational Linguistics.

\bibitem[{Kotonya and
  Toni(2020{\natexlab{b}})}]{kotonya-toni-2020-explainable-automated}
Neema Kotonya and Francesca Toni. 2020{\natexlab{b}}.
\newblock \href {https://doi.org/10.18653/v1/2020.emnlp-main.623} {Explainable
  automated fact-checking for public health claims}.
\newblock In \emph{Proceedings of the 2020 Conference on Empirical Methods in
  Natural Language Processing (EMNLP)}, pages 7740--7754, Online. Association
  for Computational Linguistics.

\bibitem[{Lee et~al.(2021)Lee, Won, Kim, Lee, Park, and Jung}]{lee2021crossaug}
Minwoo Lee, Seungpil Won, Juae Kim, Hwanhee Lee, Cheoneum Park, and Kyomin
  Jung. 2021.
\newblock Crossaug: A contrastive data augmentation method for debiasing fact
  verification models.
\newblock In \emph{Proceedings of the 30th ACM International Conference on
  Information \& Knowledge Management}, CIKM '21. Association for Computing
  Machinery.

\bibitem[{Lehmann et~al.(2015)Lehmann, Isele, Jakob, Jentzsch, Kontokostas,
  Mendes, Hellmann, Morsey, Van~Kleef, Auer et~al.}]{lehmann2015dbpedia}
Jens Lehmann, Robert Isele, Max Jakob, Anja Jentzsch, Dimitris Kontokostas,
  Pablo~N Mendes, Sebastian Hellmann, Mohamed Morsey, Patrick Van~Kleef,
  S{\"o}ren Auer, et~al. 2015.
\newblock Dbpedia--a large-scale, multilingual knowledge base extracted from
  wikipedia.
\newblock \emph{Semantic web}, 6(2):167--195.

\bibitem[{Liang et~al.(2022)Liang, Meng, Liu, Liu, Tu, Wang, Zhou, Liu, and
  Sun}]{liang2022kgsurvey}
Ke~Liang, Lingyuan Meng, Meng Liu, Yue Liu, Wenxuan Tu, Siwei Wang, Sihang
  Zhou, Xinwang Liu, and Fuchun Sun. 2022.
\newblock \href {https://doi.org/10.48550/ARXIV.2212.05767} {Reasoning over
  different types of knowledge graphs: Static, temporal and multi-modal}.

\bibitem[{Liu et~al.(2019)Liu, Ott, Goyal, Du, Joshi, Chen, Levy, Lewis,
  Zettlemoyer, and Stoyanov}]{DBLP:journals/corr/abs-1907-11692}
Yinhan Liu, Myle Ott, Naman Goyal, Jingfei Du, Mandar Joshi, Danqi Chen, Omer
  Levy, Mike Lewis, Luke Zettlemoyer, and Veselin Stoyanov. 2019.
\newblock \href {http://arxiv.org/abs/1907.11692} {Roberta: {A} robustly
  optimized {BERT} pretraining approach}.
\newblock \emph{CoRR}, abs/1907.11692.

\bibitem[{Miller(1995)}]{miller1995wordnet}
George~A Miller. 1995.
\newblock Wordnet: a lexical database for english.
\newblock \emph{Communications of the ACM}, 38(11):39--41.

\bibitem[{Park et~al.(2021)Park, Min, Kang, Zettlemoyer, and
  Hajishirzi}]{park2021faviq}
Jungsoo Park, Sewon Min, Jaewoo Kang, Luke Zettlemoyer, and Hannaneh
  Hajishirzi. 2021.
\newblock Faviq: Fact verification from information-seeking questions.
\newblock \emph{arXiv preprint arXiv:2107.02153}.

\bibitem[{Peng et~al.(2019)Peng, Yan, and Lu}]{peng2019transfer}
Yifan Peng, Shankai Yan, and Zhiyong Lu. 2019.
\newblock Transfer learning in biomedical natural language processing: An
  evaluation of bert and elmo on ten benchmarking datasets.
\newblock In \emph{Proceedings of the 2019 Workshop on Biomedical Natural
  Language Processing (BioNLP 2019)}, pages 58--65.

\bibitem[{Petroni et~al.(2019)Petroni, Rockt{\"a}schel, Riedel, Lewis, Bakhtin,
  Wu, and Miller}]{petroni-etal-2019-language}
Fabio Petroni, Tim Rockt{\"a}schel, Sebastian Riedel, Patrick Lewis, Anton
  Bakhtin, Yuxiang Wu, and Alexander Miller. 2019.
\newblock \href {https://doi.org/10.18653/v1/D19-1250} {Language models as
  knowledge bases?}
\newblock In \emph{Proceedings of the 2019 Conference on Empirical Methods in
  Natural Language Processing and the 9th International Joint Conference on
  Natural Language Processing (EMNLP-IJCNLP)}, pages 2463--2473, Hong Kong,
  China. Association for Computational Linguistics.

\bibitem[{Qi et~al.(2020)Qi, Zhang, Zhang, Bolton, and Manning}]{qi2020stanza}
Peng Qi, Yuhao Zhang, Yuhui Zhang, Jason Bolton, and Christopher~D. Manning.
  2020.
\newblock \href {https://nlp.stanford.edu/pubs/qi2020stanza.pdf} {Stanza: A
  {Python} natural language processing toolkit for many human languages}.
\newblock In \emph{Proceedings of the 58th Annual Meeting of the Association
  for Computational Linguistics: System Demonstrations}.

\bibitem[{Radford et~al.(2019)Radford, Wu, Child, Luan, Amodei, and
  Sutskever}]{radford2019language}
Alec Radford, Jeff Wu, Rewon Child, David Luan, Dario Amodei, and Ilya
  Sutskever. 2019.
\newblock Language models are unsupervised multitask learners.

\bibitem[{Raffel et~al.(2022)Raffel, Shazeer, Roberts, Lee, Narang, Matena,
  Zhou, Li, and Liu}]{raffel2019t5}
Colin Raffel, Noam Shazeer, Adam Roberts, Katherine Lee, Sharan Narang, Michael
  Matena, Yanqi Zhou, Wei Li, and Peter~J. Liu. 2022.
\newblock Exploring the limits of transfer learning with a unified text-to-text
  transformer.
\newblock \emph{J. Mach. Learn. Res.}, 21(1).

\bibitem[{Sakaguchi et~al.(2019)Sakaguchi, Bras, Bhagavatula, and
  Choi}]{sakaguchi2019winogrande}
Keisuke Sakaguchi, Ronan~Le Bras, Chandra Bhagavatula, and Yejin Choi. 2019.
\newblock Winogrande: An adversarial winograd schema challenge at scale.
\newblock \emph{arXiv preprint arXiv:1907.10641}.

\bibitem[{Sathe et~al.(2020)Sathe, Ather, Le, Perry, and
  Park}]{wikifactchkeng:2020:LREC}
Aalok Sathe, Salar Ather, Tuan~Manh Le, Nathan Perry, and Joonsuk Park. 2020.
\newblock \href {https://www.aclweb.org/anthology/2020.lrec-1.849} {Automated
  fact-checking of claims from wikipedia}.
\newblock In \emph{Proceedings of The 12th Language Resources and Evaluation
  Conference}, pages 6874--6882, Marseille, France. European Language Resources
  Association.

\bibitem[{Schuster et~al.(2021)Schuster, Fisch, and
  Barzilay}]{schuster-etal-2021-get}
Tal Schuster, Adam Fisch, and Regina Barzilay. 2021.
\newblock \href {https://doi.org/10.18653/v1/2021.naacl-main.52} {Get your
  vitamin {C}! robust fact verification with contrastive evidence}.
\newblock In \emph{Proceedings of the 2021 Conference of the North American
  Chapter of the Association for Computational Linguistics: Human Language
  Technologies}, pages 624--643, Online. Association for Computational
  Linguistics.

\bibitem[{Schuster et~al.(2019)Schuster, Shah, Yeo, Roberto Filizzola~Ortiz,
  Santus, and Barzilay}]{2019-towards-debiasing}
Tal Schuster, Darsh Shah, Yun Jie~Serene Yeo, Daniel Roberto Filizzola~Ortiz,
  Enrico Santus, and Regina Barzilay. 2019.
\newblock \href {https://doi.org/10.18653/v1/D19-1341} {Towards debiasing fact
  verification models}.
\newblock In \emph{Proceedings of the 2019 Conference on Empirical Methods in
  Natural Language Processing and the 9th International Joint Conference on
  Natural Language Processing (EMNLP-IJCNLP)}, pages 3419--3425, Hong Kong,
  China. Association for Computational Linguistics.

\bibitem[{Thorne and Vlachos(2021)}]{thorne-vlachos-2021-elastic}
James Thorne and Andreas Vlachos. 2021.
\newblock \href {https://doi.org/10.18653/v1/2021.eacl-main.82} {Elastic weight
  consolidation for better bias inoculation}.
\newblock In \emph{Proceedings of the 16th Conference of the European Chapter
  of the Association for Computational Linguistics: Main Volume}, pages
  957--964, Online. Association for Computational Linguistics.

\bibitem[{Thorne et~al.(2018)Thorne, Vlachos, Christodoulopoulos, and
  Mittal}]{thorne-etal-2018-fever}
James Thorne, Andreas Vlachos, Christos Christodoulopoulos, and Arpit Mittal.
  2018.
\newblock \href {https://doi.org/10.18653/v1/N18-1074} {{FEVER}: a large-scale
  dataset for fact extraction and {VER}ification}.
\newblock In \emph{Proceedings of the 2018 Conference of the North {A}merican
  Chapter of the Association for Computational Linguistics: Human Language
  Technologies, Volume 1 (Long Papers)}, pages 809--819, New Orleans,
  Louisiana. Association for Computational Linguistics.

\bibitem[{Toutanova and Chen(2015)}]{2015-observed}
Kristina Toutanova and Danqi Chen. 2015.
\newblock \href {https://doi.org/10.18653/v1/W15-4007} {Observed versus latent
  features for knowledge base and text inference}.
\newblock In \emph{Proceedings of the 3rd Workshop on Continuous Vector Space
  Models and their Compositionality}, pages 57--66, Beijing, China. Association
  for Computational Linguistics.

\bibitem[{Wang and Komatsuzaki(2021)}]{gpt-j}
Ben Wang and Aran Komatsuzaki. 2021.
\newblock {GPT-J-6B: A 6 Billion Parameter Autoregressive Language Model}.
\newblock \url{https://github.com/kingoflolz/mesh-transformer-jax}.

\bibitem[{Wang et~al.(2021)Wang, Mahajan, Danilevsky, and
  Rosenthal}]{wang2021semeval}
Nancy~XR Wang, Diwakar Mahajan, Marina Danilevsky, and Sara Rosenthal. 2021.
\newblock Semeval-2021 task 9: Fact verification and evidence finding for
  tabular data in scientific documents (sem-tab-facts).
\newblock \emph{arXiv preprint arXiv:2105.13995}.

\bibitem[{Williams et~al.(2018)Williams, Nangia, and Bowman}]{N18-1101}
Adina Williams, Nikita Nangia, and Samuel Bowman. 2018.
\newblock \href {http://aclweb.org/anthology/N18-1101} {A broad-coverage
  challenge corpus for sentence understanding through inference}.
\newblock In \emph{Proceedings of the 2018 Conference of the North American
  Chapter of the Association for Computational Linguistics: Human Language
  Technologies, Volume 1 (Long Papers)}, pages 1112--1122. Association for
  Computational Linguistics.

\bibitem[{Yule and Widdowson(1996)}]{yule1996pragmatics}
G.~Yule and H.G. Widdowson. 1996.
\newblock \href {https://books.google.co.kr/books?id=E2SA8ao0yMAC}
  {\emph{Pragmatics}}.
\newblock Oxford Introduction to Language Study ELT. OUP Oxford.

\bibitem[{Zhou et~al.(2019)Zhou, Han, Yang, Liu, Wang, Li, and
  Sun}]{zhou-etal-2019-gear}
Jie Zhou, Xu~Han, Cheng Yang, Zhiyuan Liu, Lifeng Wang, Changcheng Li, and
  Maosong Sun. 2019.
\newblock \href {https://doi.org/10.18653/v1/P19-1085} {{GEAR}: Graph-based
  evidence aggregating and reasoning for fact verification}.
\newblock In \emph{Proceedings of the 57th Annual Meeting of the Association
  for Computational Linguistics}, pages 892--901, Florence, Italy. Association
  for Computational Linguistics.

\end{thebibliography}
\bibliographystyle{acl_natbib}

\clearpage
\appendix

\section{Qualitative analysis}
We report claims and the retrieved graphical evidence in Table~\ref{app:qual}. 
We also report the correctness of the prediction of GEAR at the first column of our table, \textbf{Result}. 
We used subgraph retrieval to retrieve graph path visualize one of them. 
By checking the retrieved evidence, We can recognize why the model verdict the claims as refuted or supported. This shows that our graph evidence is fully interpretable.
\begin{table*}[t]
\centering
\resizebox{\linewidth}{!}{%
\begin{tabular}{l|l|l}
\Xhline{1.5pt} 
\textbf{Result}  & \textbf{Claim} & \textbf{Retrieved Path}\\ 
\hline
 \multirow{2}{*}{\textit{Correct}} & Yeah! Alfredo Zitarrosa died in a city in Uruguay & (Uruguay, ~country, Montevideo, ~deathPlace, Alfredo\_Zitarrosa) \\
 & I have heard that Mobyland had a successor. & (Mobyland, successor, ``Aero 2'')\\
 \hline
 \textit{Wrong} & I realized that a book was written by J. V. Jones and has the OCLC number 51969173 & (J.\_V.\_Jones, ~author, A\_Cavern\_of\_Black\_Ice, `oclc', ``39456030'')\\
 \Xhline{1.5pt} 
\end{tabular}
}
\caption{
    Examples of claims in \textsc{FactKG} and retrieved graph path.
}
\label{tab:qual}
\end{table*} 

\label{app:qual}

\section{Relation Substitution}
\label{app:relsub}
\begin{table*}[h]
\centering
\resizebox{0.9\textwidth}{!}{%
\begin{tabular}{llll}
\hline
\textbf{Group number} & \textbf{Head type} & \textbf{Tail type} & \textbf{Relation set} \\ \hline
\multirow{2}{*}{1} & \multirow{2}{*}{person} & \multirow{2}{*}{person} & {[}child, children{]}, {[}successor{]}, {[}parent{]}, {[}predecessor, precededBy{]}, \\
 &  &  & {[}spouse{]}, {[}vicePresident, vicepresident{]}, {[}primeminister, primeMinister{]} \\ \hline
2 & person & team & {[}currentteam, currentclub, team{]}, {[}debutTeam, formerTeam{]} \\ \hline
\multirow{3}{*}{3} & \multirow{3}{*}{non-person} & \multirow{3}{*}{person} & {[}chairperson, chairman, leader, leaderName{]}, {[}manager{]}, {[}founder{]}, \\
 &  &  & {[}director{]}, {[}crewMembers{]}, {[}producer{]},  {[}discoverer{]}, {[}creator{]}, {[}editor{]}, \\
 &  &  & {[}writer{]}, {[}coach{]}, {[}starring{]}, {[}dean{]} \\ \hline
4 & non-person & non-person & {[}owningCompany, parentCompany, owner{]},  {[}headquarter{]}, {[}builder{]} \\ \hline
\end{tabular}%
}
\caption{Group information of \textit{Relation Substitution}. 
\label{tab:apprelsub}}
\end{table*}
The four groups of compatible relations are listed in Table \ref{tab:apprelsub}.

\section{Full List of Templates}
\label{app:temp}

\subsection{Existence}
\label{app:relx}

\begin{table*}[h]
\resizebox{\textwidth}{!}{%
\begin{tabular}{m{2.2cm} m{7cm} m{6cm} m{5.5cm}}
\hline
\textbf{Type} & \textbf{Relation} & \textbf{Template} & \textbf{Example sentences} \\ \hline
\multirow{6}{*}{\textbf{Head-Relation}} & \multirow{2}{*}{\begin{tabular}[c]{@{}l@{}}successor, spouse, children, parentCompany,  \\ capital, garrison, nickname, mascot,\end{tabular}} & \multirow{2}{*}{\{Head\} had a(an) \{Relation\}.} & \multirow{2}{*}{Obama had a spouse.} \\ 
 &  &  &  \\ \cline{3-4} 
 & \multirow{2}{*}{\begin{tabular}[c]{@{}l@{}}youthclubs, predecessor, child, precededBy, \\ religion, awards, award\end{tabular}} & \multirow{2}{*}{\{Head\} did not have a(an) \{Relation\}.} & \multirow{2}{*}{Apple did not have a parent company.} \\
 &  &  &  \\ \cline{2-4} 
 & college, university & \{Head\} attended \{Relation\}. &  Obama attended university.\\ \cline{3-4} 
 &  & \{Head\} did not attend \{Relation\}. & Obama did not attend college. \\ \hline
\multirow{2}{*}{\textbf{Tail-Relation}} & \multirow{2}{*}{\begin{tabular}[c]{@{}l@{}}president, primeMinister, vicepresident,\\ primeminister,  vicePresident\end{tabular}} & \{Tail\} was a \{Relation\}. &  Obama was a president.\\ \cline{3-4} 
 &  & \{Tail\} was not a \{Relation\}. &  Obama was not a vice president.\\ \hline
\end{tabular}%
}
\caption{Templates for Existence claims. 
\label{tab:apprel}}
\end{table*}

The templates to generate existence claims are described in Table \ref{tab:apprel}.

\subsection{Factive and Non Factive Presupposition}
\label{app:factpre}
Factive and Non Factive presupposition templates are in Table \ref{tab:factpresup}.
\begin{table*}[h]
\centering
\resizebox{0.8\textwidth}{!}{%
\begin{tabular}{lll}
\hline
\textbf{Presupposition type} & \textbf{Template} & \textbf{Claim Example}\\ \hline
\multirow{8}{*}{\textbf{Factive}} & I forgot that \{claim\}. & I forgot that Obama was president. \\ \cline{2-3} 
 & I realized that \{claim\}. & I realized that Obama was president. \\ \cline{2-3} 
 & I wasn’t aware that \{claim\}. & I wasn’t aware that Obama was president. \\ \cline{2-3} 
 & I didn’t know that \{claim\}. & I didn’t know that Obama was president. \\ \cline{2-3} 
 & I remembered that \{claim\}. & I remembered that Obama was president. \\ \cline{2-3} 
 & I explained that \{claim\}. & I explained that Obama was president. \\ \cline{2-3} 
 & I emphasized that \{claim\}. & I emphasized that Obama was president. \\ \cline{2-3} 
 & I understand that \{claim\}. & I understand that Obama was president. \\ \hline
\multirow{3}{*}{\textbf{Non Factive}} & I imagined that \{claim\}. & I imagined that Obama was president. \\ \cline{2-3} 
 & I wish that \{claim\}. & I wish that Obama was president. \\ \cline{2-3} 
 & If only \{claim\}. & If only Obama was president. \\ \hline
\end{tabular}%
}
\caption{Templates for factive, non factive presupposition.
\label{tab:factpresup}
}
\end{table*}

\subsection{Structural Presupposition}
\label{app:structpre}
Structural presupposition templates are in  Table \ref{tab:strcutprep}.
\begin{table*}[]
\resizebox{\textwidth}{!}{%
\begin{tabular}{lllll}
\hline
\multicolumn{2}{l}{\textbf{Type}} & \textbf{Relations} & \textbf{Template} & \textbf{Example claim} \\ \hline
\multicolumn{2}{l}{\multirow{30}{*}{\textbf{One-hop}}} & leader, leaderName, mayor, & \multirow{4}{*}{When was \{tail\} a \{relaion\} of \{head\}?} & \multirow{4}{*}{When was Elizabeth II a leader of Alderney?} \\
\multicolumn{2}{l}{} & senators, president, manager, &  &  \\
\multicolumn{2}{l}{} & generalManager, coach, &  &  \\
\multicolumn{2}{l}{} & chairman, dean &  &  \\ \cline{3-5} 
\multicolumn{2}{l}{} & team, draftTeam, clubs, & \multirow{2}{*}{When did \{head\} play for \{tail\}?} & \multirow{2}{*}{\begin{tabular}[c]{@{}l@{}}When did Aaron Boogaard play for Wichita \\ Thunder?\end{tabular}} \\
\multicolumn{2}{l}{} & managerClub, managerclubs &  &  \\ \cline{3-5} 
\multicolumn{2}{l}{} & operator & When did \{tail\} operate \{head\}? & When did Aktieselskab operate Aarhus Airport? \\ \cline{3-5} 
\multicolumn{2}{l}{} & occupation, formerName & When was \{head\} a \{tail\}? & When was HBO a The Green Channel? \\ \cline{3-5} 
\multicolumn{2}{l}{} & \multirow{2}{*}{almaMater} & \multirow{2}{*}{When did \{head\} graduate from the \{tail\}?} & \multirow{2}{*}{\begin{tabular}[c]{@{}l@{}}When did Ab Klink graduate from the Erasmus \\ University Rotterdam?\end{tabular}} \\
\multicolumn{2}{l}{} &  &  &  \\ \cline{3-5} 
\multicolumn{2}{l}{} & fossil & When was \{tail\} fossil found in \{head\}? & When was Smilodon fossil found in California? \\ \cline{3-5} 
\multicolumn{2}{l}{} & \multirow{2}{*}{director} & \multirow{2}{*}{When was \{head\} directed by \{tail\}?} & \multirow{2}{*}{\begin{tabular}[c]{@{}l@{}}When was Death on a Factory Farm directed by \\ Sarah Teale?\end{tabular}} \\
\multicolumn{2}{l}{} &  &  &  \\ \cline{3-5} 
\multicolumn{2}{l}{} & \multirow{2}{*}{producer} & \multirow{2}{*}{When was \{head\} produced by \{tail\}?} & \multirow{2}{*}{\begin{tabular}[c]{@{}l@{}}When was Turn Me On (album) produced by \\ Wharton Tiers?\end{tabular}} \\
\multicolumn{2}{l}{} &  &  &  \\ \cline{3-5} 
\multicolumn{2}{l}{} & \multirow{2}{*}{\begin{tabular}[c]{@{}l@{}}foundation, foundedBy, \\ founder\end{tabular}} & \multirow{2}{*}{When was \{head\} founded by \{tail\}?} & \multirow{2}{*}{\begin{tabular}[c]{@{}l@{}}When was MotorSport Vision founded by \\ Jonathan Palmer?\end{tabular}} \\
\multicolumn{2}{l}{} &  &  &  \\ \cline{3-5} 
\multicolumn{2}{l}{} & deathCause & When did \{head\} die from \{tail\}? & When did James Craig Watson die from Peritonitis? \\ \cline{3-5} 
\multicolumn{2}{l}{} & creators, creator & When was \{head\} created by \{tail\}? & When was April O'Neil created by Peter Laird? \\ \cline{3-5} 
\multicolumn{2}{l}{} & starring & When was \{head\} starring \{tail\}? & When was Bananaman starring Graeme Garden? \\ \cline{3-5} 
\multicolumn{2}{l}{} & shipBuilder, builder & When was \{head\} built by \{tail\}? & When was A-Rosa Luna built by Germany? \\ \cline{3-5} 
\multicolumn{2}{l}{} & \multirow{2}{*}{designer} & \multirow{2}{*}{When was \{head\} designed by \{tail\}?} & \multirow{2}{*}{\begin{tabular}[c]{@{}l@{}}When was Atatürk Monument (İzmir) designed by \\ Pietro Canonica?\end{tabular}} \\
\multicolumn{2}{l}{} &  &  &  \\ \cline{3-5} 
\multicolumn{2}{l}{} & \multirow{2}{*}{shipCountry} & \multirow{2}{*}{When did \{head\} come from \{tail\}?} & \multirow{2}{*}{\begin{tabular}[c]{@{}l@{}}When did ARA Veinticinco de Mayo (V-2) come \\ from Argentina?\end{tabular}} \\
\multicolumn{2}{l}{} &  &  &  \\ \cline{3-5} 
\multicolumn{2}{l}{} & \multirow{2}{*}{spouse} & \multirow{2}{*}{When was \{head\} married to \{tail\}?} & \multirow{2}{*}{\begin{tabular}[c]{@{}l@{}}When was Abraham A. Ribicoff married to Ruth \\ Ribicoff?\end{tabular}} \\
\multicolumn{2}{l}{} &  &  &  \\ \cline{3-5} 
\multicolumn{2}{l}{} & champions & When was \{tail\} champion at the \{head\}? & When was Juventus F.C. champion at the Serie A? \\ \cline{3-5} 
\multicolumn{2}{l}{} & \multirow{2}{*}{recordedIn} & \multirow{2}{*}{When was \{head\} recorded in \{tail\}?} & \multirow{2}{*}{\begin{tabular}[c]{@{}l@{}}When was Bootleg Series Volume 1: The Quine Tapes \\ recorded in San Francisco?\end{tabular}} \\
\multicolumn{2}{l}{} &  &  &  \\ \hline
\multirow{10}{*}{\textbf{Existence}} & \multirow{7}{*}{\textbf{Head-Relation}} & successor, spouse, children, & \multirow{6}{*}{What is the name of \{head\}'s  \{relation\}?} & \multirow{6}{*}{What is the name of Obama's child?} \\
 &  & parentCompany, capital, &  &  \\
 &  & garrison, nickname, mascot, &  &  \\
 &  & youthclubs, predecessor, &  &  \\
 &  & child, precededBy, religion, &  &  \\
 &  & awards, award &  &  \\ \cline{3-5} 
 &  & college, university & When did \{head\} attend \{relation\}? & When did Obama attend university? \\ \cline{2-5} 
 & \multirow{3}{*}{\textbf{Tail-Relation}} & president, primeMinister, & When was \{tail\} \{relation\}? & When was Obama President? \\ \cline{4-5} 
 &  & vicepresident, primeminister, & Where was \{tail\} \{relation\}? & Where was Biden Vice President? \\ \cline{4-5} 
 &  & vicePresident & What country was \{tail\} \{relation\}? & What country was Obama President? \\ \hline
\end{tabular}%
}
\caption{Templates for structural presupposition.
\label{tab:strcutprep}
}
\end{table*}

\section{Negation Labeling}

\subsection{Conjunction}
\label{app:negconj}
\begin{table*}[]
\resizebox{\textwidth}{!}{%
\begin{tabular}{m{4cm}|m{10cm}| >{\centering\arraybackslash}m{3cm}}
\Xhline{1.5pt}

\multicolumn{1}{c|}{\textbf{Graph}} & \multicolumn{1}{c|}{\textbf{Claim Example}} & \textbf{Label} \\ \Xhline{1pt}

\begin{tikzpicture}[node distance={15mm}, main/.style = {draw, circle}] 
\node[main] (1){$a$};
\node[main] (2) [right of=1] {$m$}; 
\node[main] (3) [right of=2] {$p$};
\draw[->] (1) -- node[midway, above] {$r_2$} (2);
\draw[->] (2) -- node[midway, above] {$r_4$} (3);
\end{tikzpicture}
&  AIDAstella was built by Meyer Werft in Papenburg.&
\textsc{Supported} \\ \hline

\begin{tikzpicture}[node distance={15mm}, main/.style = {draw, circle}] 
\node[main] (1){$a$};
\node[main] (2) [right of=1] {$m$}; 
\node[main] (3) [right of=2, red] {$n$};
\draw[->] (1) -- node[midway, above] {$r_2$} (2);
\draw[->] (2) -- node[midway, above] {$r_4$} (3);
\end{tikzpicture}
&  AIDAstella was built by Meyer Werft in \textcolor{red}{New York}.&
\textsc{Refuted}  \\ \hline

\begin{tikzpicture}[node distance={15mm}, main/.style = {draw, circle}] 
\node[main] (1){$a$};
\node[main] (2) [right of=1] {$m$}; 
\node[main] (3) [right of=2, red] {$n$};
\draw[->, red] (1) -- node[midway, above] {$r_2$} (2);
\draw[->] (2) -- node[midway, above] {$r_4$} (3);
\end{tikzpicture}
&  AIDAstella was \textcolor{red}{not} built by Meyer Werft in \textcolor{red}{New York}.&
\textsc{Refuted}  \\ \hline

\begin{tikzpicture}[node distance={15mm}, main/.style = {draw, circle}] 
\node[main] (1){$a$};
\node[main] (2) [right of=1] {$m$}; 
\node[main] (3) [right of=2, red] {$n$};
\draw[->] (1) -- node[midway, above] {$r_2$} (2);
\draw[->, red] (2) -- node[midway, above] {$r_4$} (3);
\end{tikzpicture}
&  AIDAstella was built by Meyer Werft, \textcolor{red}{not} in \textcolor{red}{New York}.&
\textsc{Supported}  \\ \hline

\begin{tikzpicture}[node distance={15mm}, main/.style = {draw, circle}] 
\node[main] (1){$a$};
\node[main] (2) [right of=1] {$m$}; 
\node[main] (3) [right of=2, red] {$n$};
\draw[->, red] (1) -- node[midway, above] {$r_2$} (2);
\draw[->, red] (2) -- node[midway, above] {$r_4$} (3);
\end{tikzpicture}
&  AIDAstella was \textcolor{red}{not} built by Meyer Werft, \textcolor{red}{not} in \textcolor{red}{New York}.&
\textsc{Refuted}  \\ \Xhline{1.5pt}
\end{tabular}%
}
\caption{$r_2$: shipBuilder, $r_4$: location, $m$: Meyer Werft, $a$: AIDAstella, $n$: New York, $p$: Papenburg. 
\label{tab:negconjr}
}
\end{table*}
\begin{table*}[]
\resizebox{\textwidth}{!}{%
\begin{tabular}{m{4cm}|m{10cm}| >{\centering\arraybackslash}m{3cm}}
\Xhline{1.5pt}

\multicolumn{1}{c|}{\textbf{Graph}} & \multicolumn{1}{c|}{\textbf{Claim Example}} & \textbf{Label} \\ \Xhline{1pt}

\begin{tikzpicture}[node distance={15mm}, main/.style = {draw, circle}] 
\node[main] (1){$a$};
\node[main] (2) [right of=1] {$m$}; 
\node[main] (3) [right of=2] {$p$};
\draw[->] (1) -- node[midway, above] {$r_2$} (2);
\draw[->] (2) -- node[midway, above] {$r_4$} (3);
\end{tikzpicture}
&  AIDAstella was built by Meyer Werft in Papenburg.&
\textsc{Supported} \\ \hline

\begin{tikzpicture}[node distance={15mm}, main/.style = {draw, circle}] 
\node[main] (1){$a$};
\node[main] (2) [right of=1, red] {$s$}; 
\node[main] (3) [right of=2] {$p$};
\draw[->] (1) -- node[midway, above] {$r_2$} (2);
\draw[->] (2) -- node[midway, above] {$r_4$} (3);
\end{tikzpicture}
&  AIDAstella was built by \textcolor{red}{Samsung} in Papenburg.&
\textsc{Refuted}  \\ \hline

\begin{tikzpicture}[node distance={15mm}, main/.style = {draw, circle}] 
\node[main] (1){$a$};
\node[main] (2) [right of=1, red] {$s$}; 
\node[main] (3) [right of=2] {$p$};
\draw[->, red] (1) -- node[midway, above] {$r_2$} (2);
\draw[->] (2) -- node[midway, above] {$r_4$} (3);
\end{tikzpicture}
&  AIDAstella was \textcolor{red}{not} built by \textcolor{red}{Samsung} in Papenburg.&
\textsc{Refuted}  \\ \hline

\begin{tikzpicture}[node distance={15mm}, main/.style = {draw, circle}] 
\node[main] (1){$a$};
\node[main] (2) [right of=1, red] {$s$}; 
\node[main] (3) [right of=2] {$p$};
\draw[->] (1) -- node[midway, above] {$r_2$} (2);
\draw[->, red] (2) -- node[midway, above] {$r_4$} (3);
\end{tikzpicture}
&  AIDAstella was built by \textcolor{red}{Samsung}, \textcolor{red}{not} in Papenburg.&
\textsc{Refuted}  \\ \hline

\begin{tikzpicture}[node distance={15mm}, main/.style = {draw, circle}] 
\node[main] (1){$a$};
\node[main] (2) [right of=1, red] {$s$}; 
\node[main] (3) [right of=2] {$p$};
\draw[->, red] (1) -- node[midway, above] {$r_2$} (2);
\draw[->, red] (2) -- node[midway, above] {$r_4$} (3);
\end{tikzpicture}
&  AIDAstella was \textcolor{red}{not} built by \textcolor{red}{Samsung}, \textcolor{red}{not} in Papenburg.&
\textsc{Supported}  \\ \Xhline{1.5pt}
\end{tabular}%
}
\caption{$r_2$: shipBuilder, $r_4$: location, $m$: Meyer Werft, $a$: AIDAstella, $p$: Papenburg, $s$: Samsung.
\label{tab:negconjc}
}
\end{table*}
When the negation is added to \textsc{Refuted} claims, the label depends on the position of the negation. 
If negations are added to all parts with substituted entities, it becomes a \textsc{Supported} claim.
Conversely, other cases preserve the label \textsc{Refuted} since the negation is added to a place that is not related to entity substitution.
Detailed examples are described in Table~\ref{tab:negconjr} and Table~\ref{tab:negconjc}.

\subsection{Multi-hop}
\label{app:negmul}
\begin{table*}[]
\resizebox{\textwidth}{!}{%
\begin{tabular}{m{4cm}|m{10cm}| >{\centering\arraybackslash}m{3cm}}
\Xhline{1.5pt}

\multicolumn{1}{c|}{\textbf{Graph}} & \multicolumn{1}{c|}{\textbf{Claim Example}} & \textbf{Label} \\ \Xhline{1pt}

\begin{tikzpicture}[node distance={15mm}, main/.style = {draw, circle}] 
\node[main] (1){$s$};
\node[main] (2) [right of=1, dashed] {$x$}; 
\node[main] (3) [right of=2] {$p$};
\draw[->] (1) -- node[midway, above] {$r_2$} (2);
\draw[->] (2) -- node[midway, above] {$r_4$} (3);
\end{tikzpicture}
&  AIDAstella was built by a company in Papenburg.&
\textsc{Supported} \\ \hline

\begin{tikzpicture}[node distance={15mm}, main/.style = {draw, circle}] 
\node[main] (1){$s$};
\node[main] (2) [right of=1, dashed] {$x$}; 
\node[main] (3) [right of=2, red] {$n$};
\draw[->] (1) -- node[midway, above] {$r_2$} (2);
\draw[->] (2) -- node[midway, above] {$r_4$} (3);
\end{tikzpicture}
&  AIDAstella was built by a company in \textcolor{red}{New York}.&
\textsc{Refuted}  \\ \hline

\begin{tikzpicture}[node distance={15mm}, main/.style = {draw, circle}] 
\node[main] (1){$s$};
\node[main] (2) [right of=1, dashed] {$x$}; 
\node[main] (3) [right of=2, red] {$n$};
\draw[->, red] (1) -- node[midway, above] {$r_2$} (2);
\draw[->] (2) -- node[midway, above] {$r_4$} (3);
\end{tikzpicture}
&  AIDAstella was \textcolor{red}{not} built by a company in \textcolor{red}{New York}.&
\textsc{Path Check}  \\ \hline

\begin{tikzpicture}[node distance={15mm}, main/.style = {draw, circle}] 
\node[main] (1){$s$};
\node[main] (2) [right of=1, dashed] {$x$}; 
\node[main] (3) [right of=2, red] {$n$};
\draw[->] (1) -- node[midway, above] {$r_2$} (2);
\draw[->, red] (2) -- node[midway, above] {$r_4$} (3);
\end{tikzpicture}
&  AIDAstella was built by a company, \textcolor{red}{not} in \textcolor{red}{New York}.&
\textsc{Path Check}  \\ \hline

\begin{tikzpicture}[node distance={15mm}, main/.style = {draw, circle}] 
\node[main] (1){$s$};
\node[main] (2) [right of=1, dashed] {$x$}; 
\node[main] (3) [right of=2, red] {$n$};
\draw[->, red] (1) -- node[midway, above] {$r_2$} (2);
\draw[->, red] (2) -- node[midway, above] {$r_4$} (3);
\end{tikzpicture}
&  AIDAstella was \textcolor{red}{not} built by a company, \textcolor{red}{not} in \textcolor{red}{New York}.&
\textsc{Path Check}  \\ \Xhline{1.5pt}
\end{tabular}%
}
\caption{$r_2$: shipBuilder, $r_4$: location, $s$: AIDAstella, $n$: New York. 
\label{tab:negmulti}
}
\end{table*}
The truth of this claim is dependent on the presence of a distinctive path that matches the claim's intent.
For example, when verifying the claim in the fourth row of the Table~\ref{tab:negmulti}, we check the existence of an entity which is connected to \textit{`AIDAstella'} with relation \textit{builder} and not connected to \textit{`New York'} with relation \textit{location}.

\section{Colloquial Style Claim Survey}
\label{app:survey}
A total of 9 graduate students participated in the survey to evaluate how much information was lost  in the colloquial style claim compared to original claim. Since each person has different criteria for `important information', the labels are divided into five rather than two.
The labels are as follows, 
i) All facts are preserved,
ii) Minor loss of information or minor grammatical errors,
iii) Ambiguous whether the lost information is important,
iv) It is ambiguous, but the lost information may be important,
v) Loss of important information.
 And as a result, only 9.8\% of the claims were selected as v) Loss of important information by at least two reviewers.

\section{Details of GEAR}
To make this paper self-contained, we recall some details of the claim verification in GEAR~\citep{zhou-etal-2019-gear}.
The authors of GEAR (\citet{zhou-etal-2019-gear}) used sentence encoder to obtain representations for the claim and the evidence. Then they built a fully-connected evidence graph and used evidence reasoning network (ERNet) to propagate information between evidence and reason over the graph. Finally, they used an evidence aggregator to infer the final results.

\subsubsection*{Sentence Encoder} 
Given an input sentence, \citet{zhou-etal-2019-gear} employed BERT~\cite{bertdevlin} as a sentence encoder by extracting the final hidden state of the [CLS] token as the representation.

Specifically, given a claim $c$ and $N$ pieces of retrieved evidence $\{ e_1, e_2, ..., e_N \}$, they fed each evidence-claim pair $(e_i, c)$ into BERT to obtain the evidence representation $\mathbf{e_i}$. they also fed the claim into BERT alone to obtain the claim $\mathbf{c}$. That is,
\begin{equation}
\begin{split}
	\mathbf{e}_i &= \text{BERT}\left(e_i, c\right), \\	
	\mathbf{c} &= \text{BERT}\left(c\right).
\end{split}
\end{equation}

\subsubsection*{Evidence Reasoning Network}
Let $\mathbf{h}^t = \{\mathbf{h}_1^t, \mathbf{h}_2^t, ..., \mathbf{h}_N^t\}$ denote the hidden states of the nodes in layer $t$, where $\mathbf{h}_i^t \in \mathcal{R}^{F \times 1}$ and $F$ is the number of features in each node. The initial hidden state of each evidence node $\mathbf{h}_i^0$ was initialized by the evidence: $\mathbf{h}_i^0 = \mathbf{e}_i$.
The authors proposed an Evidence Reasoning Network (ERNet) to propagate information among the evidence nodes. 
They first used an MLP to calculate the attention coefficients between a node $i$ and its neighbor $j$ ($j \in \mathcal{N}_i$),
\begin{equation}
	p_{ij} = \mathbf{W}_1^{t-1} ( \text{ReLU} ( \mathbf{W}_0^{t-1} ( \mathbf{h}_i^{t-1} \| \mathbf{h}_j^{t-1} ) ) ),
\end{equation}
where $\mathcal{N}_i$ denotes the set of neighbors of node $i$, $\mathbf{W}_0^{t-1} \in \mathcal{R}^{H \times 2F}$ and $\mathbf{W}_1^{t-1} \in \mathcal{R}^{1 \times H}$ are weight matrices, and $\cdot\|\cdot$ denotes the concatenation operation.

Then, they normalized the coefficients using the softmax function,
\begin{equation}
	\alpha_{ij} = \text{softmax}_j(p_{ij}) = \frac{\text{exp}(p_{ij})}{\sum_{k \in \mathcal{N}_i} \text{exp}(p_{ik})}.
\end{equation}

Finally, the normalized attention coefficients were used to compute a linear combination of the neighbor features and thus obtained the features for node $i$ at layer $t$,
\begin{equation}
	\mathbf{h}_i^t = \sum_{j \in \mathcal{N}_i} \alpha_{ij} \mathbf{h}_j^{t-1}.
\end{equation}

The authors fed the final hidden states of the evidence nodes $\{\mathbf{h}_1^T, \mathbf{h}_2^T, ...,\mathbf{h}_N^T\}$ into their evidence aggregator to make the final inference.

\subsubsection*{Evidence Aggregator}
The authors employed an evidence aggregator to gather information from different evidence nodes and obtained the final hidden state $\mathbf{o} \in \mathcal{R}^{F \times 1}$.
We used the mean aggregator in GEAR.

The mean aggregator performed the \emph{element-wise} Mean operation among hidden states.
	\begin{equation}
		\mathbf{o} = \text{Mean} (\mathbf{h}_1^T, \mathbf{h}_2^T, ..., \mathbf{h}_N^T).
	\end{equation}
	
Once the final state $\mathbf{o}$ is obtained, the authors employed a one-layer MLP to get the final prediction $l$.
\begin{equation}
	l = \text{softmax} (\text{ReLU}( \mathbf{W} \mathbf{o} + \mathbf{b})),
\end{equation}
where $\mathbf{W} \in \mathcal{R}^{C \times F}$ and $\mathbf{b} \in \mathcal{R}^{C \times 1}$ are parameters, and $C$ is the number of prediction labels.
\label{app:gear}

\end{document}